# IMPACT OF AUTOMATIC IMAGE CLASSIFICATION AND BLIND DECONVOLUTION IN IMPROVING TEXT DETECTION PERFORMANCE OF THE CRAFT ALGORITHM

Clarisa V. Albarillo[1] and Proceso L. Fernandez, Jr.[2]

[1]College of Computer Science,
Don Mariano Marcos Memorial State University, La Union, Philippines
[2]Department of Information Systems & Computer Science,
Ateneo de Manila University, Quezon City, Philippines

## *ABSTRACT*

*Text detection in natural scenes has been a significant and active research subject in computer vision and document analysis because of its wide range of applications as evidenced by the emergence of the Robust Reading Competition. One of the algorithms which has good text detection performance in the said competition is the Character Region Awareness for Text Detection (CRAFT). Employing the ICDAR 2013 dataset, this study investigates the impact of automatic image classification and blind deconvolution as image pre-processing steps to further enhance the text detection performance of CRAFT. The proposed technique automatically classifies the scene images into two categories, blurry and non-blurry, by utilizing of a Laplacian operator with 100 as threshold. Prior to applying the CRAFT algorithm, images that are categorized as blurry are further pre-processed using blind deconvolution to reduce the blur. The results revealed that the proposed method significantly enhanced the detection performance of CRAFT, as demonstrated by its IoU h-mean of 94.47% compared to the original 91.42% h-mean of CRAFT and this even outperformed the top-ranked SenseTime, whose h-mean is 93.62%.*

## 1. INTRODUCTION

With the rise and development of deep learning, computer vision has been tremendously transformed and reshaped. A wave of change brought about by deep learning has unavoidably influenced a specific area of research in computer vision and document analysis called text detection. The community has recently experienced significant improvements in thinking, approach, and performance.

Over the years, text detection has been a popular topic in computer vision field [1, 2, 3, 4, 5] due to the wide range of applications. These applications include image search [6, 7], target geolocation [8, 9], human-computer interaction [10, 11], robot navigation [12, 13], and industrial automation [14, 15], which greatly benefits from the detailed information included in text.

Text detection still ranks highly in difficulty and is challenging because of the diversity of text patterns and complex scene structures in natural images, despite advances in the field. The diversity of text patterns and complexities of scene images include arbitrary text size and type, cluttered image background, and variation of light conditioning. Several challenges also still exist, such as noise, blur, distortion, occlusion, and variance. Recently, several text detection algorithms used the booming technology of deep learning. Scene text detectors based on deep learning have shown promising performance [16, 17, 18, 19].

The introduction of the Robust Reading Competition is an evidence that the challenge of finding texts in scene images has attracted significant interest over time. This competition was launched in 2011 and was organized with the International Conference on Document Analysis and Recognition (ICDAR). The competition is designed around a variety of complex computer vision tasks that address a wide spectrum of real-world situations.

One such challenge is the Focused Scene Text, which focuses on reading texts in real-world contexts. The scenario being examined includes "focused text"—textual images that are largely focused on the text content of interest. The Focused Scene Text Challenge has three editions: ICDAR 2011, ICDAR 2013, and ICDAR 2015. For the tasks of text localization, text segmentation, and word recognition, the ICDAR 2013 is the definitive one [20].

Additionally, the competition features three performance metrics: ICDAR 2013 evaluation, Deteval, and Intersection-over-Union (IoU), each of which has a ranking of the best algorithms. The h-means of the top-ranking algorithms for these three metrics are 96.38%, 96.78% and 93.62% respectively. These figures show that the third metric (IoU) appears to be the most difficult, hence this study is focused on improving the IoU results.

The top-performing algorithms as measured by the IoU metric are listed in Table 1. The Sensetime algorithm, with an IoU h-mean of 93.62%, is the top-ranked algorithm according to this metric. A single end-to-end trainable Fast Oriented Text Spotting (FOTS) network that is designed for simultaneous detection and recognition is used in the method. To share convolutional features across detection and identification, it specifically introduced RoIRotate [21].

Table 1. State-of-the-art results on Text Localization, based on the IoU performance metric (see Robust Reading Competition - IoU Evaluation of Task 1)

| Method | Precision | Recall | H-Mean |
|---|---|---|---|
| **Sensetime (2016)** | **91.87%** | **95.45%** | **93.62%** |
| TextFuseNet (2020) | 90.78% | 95.58% | 93.11% |
| TencentAILab (2017) | 94.79% | 91.37% | 93.05% |
| VARCO (2020) | 89.86% | 93.63% | 91.71% |
| HIT (2020) | 89.22% | 93.85% | 91.48% |
| CRAFT (2018) | 89.04% | 93.93% | 91.42% |

Majority of scene text detectors train their networks primarily to detect word level bounding boxes. Such a level could be challenging in complex situations, including texts with arbitrary font and size, different text scales and shapes, like curved or distorted fonts, or messages that are exceedingly long. In such circumstances, detectors typically return bounding boxes at the level of characters rather than the entire word. The Character Region Awareness for Text detection (CRAFT) [16], a text detector that localizes the individual character regions and then correlates the detected characters to a text instance, is one approach that may handle such situations well. It uses a convolutional neural network to generate both the affinity score, which collects all the characters into one instance, and the character region score, which is used to localize specific characters in an image.

CRAFT's IoU h-mean of 91.42% currently places the algorithm sixth in the competition. CRAFT performs well, but there is still considerable potential for improvement because it assumes that the images from ICDAR 2013 are clear and free from any blur or image distortion. A previous study [22] explored how blind deconvolution can be applied to improve text localization and recognition of texts in both clear and blurry datasets using a fast-bounding box algorithm. It employed manual classification of images into blurry and non-blurry classes. The findings of the study reveal that the performance results of the proposed text localization and recognition method using manual classification of images and blind deconvolution is significantly improved.

With this, the researchers [22] were prompted to investigate further on the impact of using Laplacian operator in automatic image classification and blind deconvolution for deblurring. This study explores on improving the detection performance of CRAFT by adding some pre-processing steps that include automatically detecting blurry images and then attempting to reduce the blur on these images prior to running the CRAFT algorithm using blind deconvolution. This proposed method is referred as BD-CRAFT, a variant of CRAFT. It also aims to show that the BD-CRAFT is not only significantly better than CRAFT, but it is also able to outperform the current best state-of-the-art algorithm, SenseTime, for scene text detection.

## 2. Methodology

### 2.1. The Dataset

The International Conference on Document Analysis and Recognition (ICDAR) 2013 Focused Scene Text Competition Challenge 2 dataset was employed in this study. Recent computer vision datasets including 2017 COCO-Text [23], deTEXT [24], DOST [25], FSNS [26], MLT [27], and IEHHR [28] were not used because the focus of this study is on scene images.

The ICDAR 2013 Challenge 2 dataset is composed of images that are specifically focused on the relevant text content to depict the use case in which a user focuses a camera on a scene text for text reading and translation applications. As such, the focus text is horizontal in most cases. Furthermore, this dataset has 229 training images, and 233 test images and provides "Strong", "Weak" and "Generic" lexicons - of different sizes for the task of text detection, similar to ICDAR 2015 [29, 30]. Moreover, the images were of various sizes and were taken in various lighting conditions and environments using different cameras. Some of the images are shown in Figure 1.

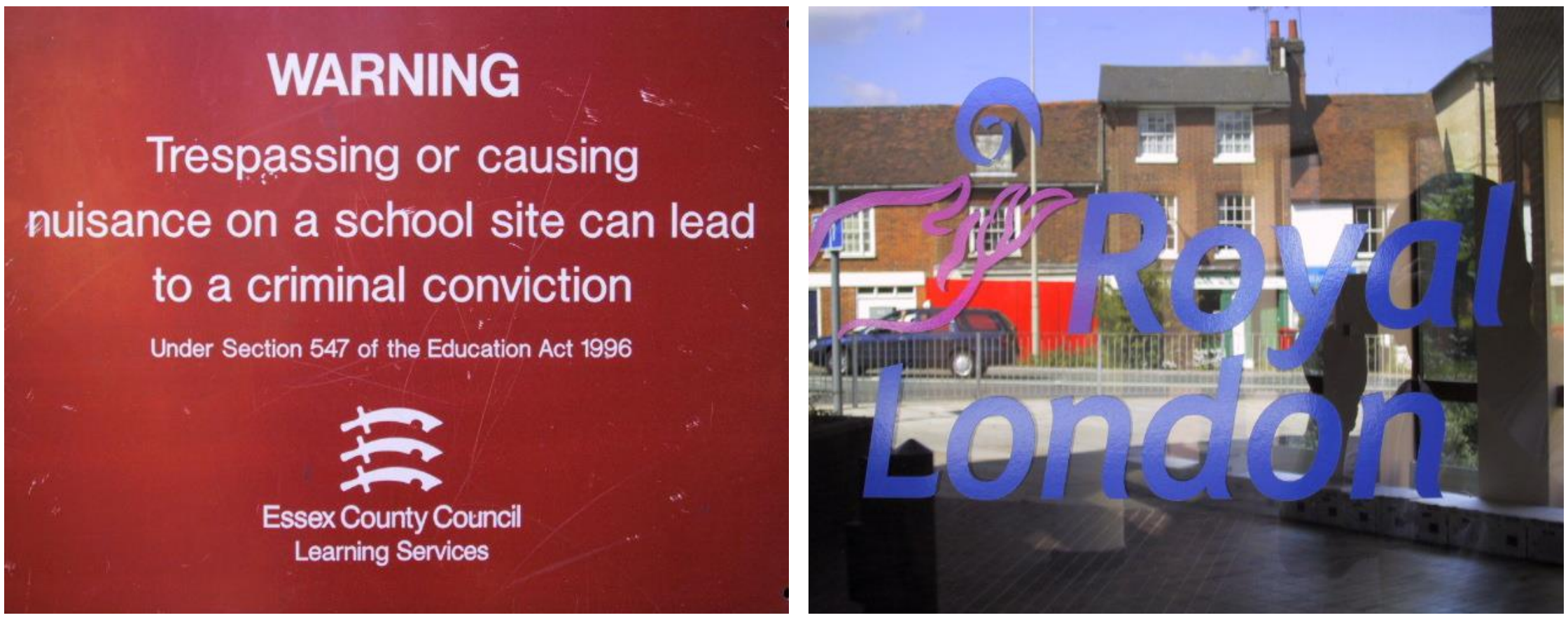

Figure 1. Sample Images from ICDAR 2013 Focused Scene Text Competition Challenge 2 Datasets

## 2.2. Exploration of the CRAFT Algorithm

To identify the weakness of the CRAFT algorithm, it was first run on the ICDAR 2013 Dataset. The CRAFT algorithm is available for download at https://pypi.org/project/craft-text-detector/. An initial goal was to attempt to identify at least one of the weaknesses of the CRAFT algorithm and then to target addressing such a weakness in order to improve the performance of CRAFT with respect to a specific metric. This specific main performance metric is detailed in the next section. It was hypothesized that this goal can be achieved by partitioning the dataset into two -- one consisting of images where CRAFT algorithm performs poorly under the metric, and the other one consisting of images where CRAFT performs well -- and then analyzing patterns common to the first partition.

## 2.3. The Evaluation Metrics

Intersection over Union (IoU), a well-known similarity metric that is evaluated as the ratio of two entities - the overlapping area and the union area - is used to assess the accuracy of the proposed method.

In the problem context, IoU quantifies how well the predicted bounding box overlaps with the ground truth box. The scale ranges from 0 (no overlap) to the ideal value of 1 (perfect overlap). An IoU value of 0.5 or higher is considered a good prediction and signifies that the texts have been located accurately [31]. In order to assess the IoU performance of the proposed algorithm, the precision, recall, and eventually the h-mean are computed from this prediction. The formulas for these well-known measures are provided below for completeness:

$$Precision = \frac{TP}{TP+FP} \quad \text{or} \quad Precision(G,D) = \frac{\sum_{j=1}^{|D|} Bestmatch_D(D_j)}{|D|}$$

$$Recall = \frac{TP}{TP+FN} \quad \text{or} \quad Recall(G,D) = \frac{\sum_{i=1}^{|G|} Bestmatch_G(G_i)}{|G|} \quad \text{and}$$

$$Hmean = 2\frac{(Recall * Precision)}{(Recall + Precision)}$$

where, $Bestmatch_D$ and $Bestmatch_G$ indicate the closest match between detection and ground truth as defined below:

$$Bestmatch_G(G_i) = \frac{2 \cdot Area(G_i \cap D_j)}{Area(G_i) + Area(D_j)}$$

$$Bestmatch_D(D_j) = \frac{2 \cdot Area(D_j \cap G_i)}{Area(D_j) + Area(G_i)}$$

Note that H-mean refers to the harmonic mean of the Precision and Recall and therefore takes into account both false positives (FP) and false negatives (FN).

## 2.4. Experimentations on Improving the Detection Performance of CRAFT

This part is the heart of the study. This impacts of automatic image classification and blind deconvolution as pre-processing steps are explored in order to improve the detection performance of CRAFT. Actually, the objective is not merely to improve CRAFT's performance, but to make the improvement sufficiently significant such that the proposed technique even outperforms all currently available state-of-the-art text detection algorithms, including the top-ranked SenseTime algorithm. This was a difficult undertaking, especially since text detection is a well-researched topic, and numerous researchers have attempted to develop algorithms that perform on the ICDAR 2013 dataset.

### 2.5. Comparison with the State-of-the-Art Algorithms

The final part of this study involves providing proof of the superiority of the proposed algorithm against existing algorithms with respect to the main performance metric. Towards this end, the performance results on the ICDAR 2013 dataset of the target algorithms are collected and compared with the result of the proposed technique. The performance results of the top-performing algorithms that were submitted to the Robust Reading Competition are published in the competition website along with the actual performance of each image.

## 3. RESULTS AND DISCUSSION

Several studies and experiments were carried out using the ICDAR 2013 dataset in an attempt to improve the performance of Character Region Awareness for Text Detection (CRAFT).

### 3.1. Identifying the Weakness of the CRAFT Algorithm

The ICDAR 2013 dataset was used as the basis for the CRAFT algorithm, which was tested against it as the first major task in the study. The dataset was divided into two categories: one containing images where the CRAFT algorithm performs poorly under the metric, and the other containing images where CRAFT performs well. This was done after confirming that the performance exactly matches what is published in the official Robust Reading Competition website [32]. Figure 2 depicts sample images from these two partitions.

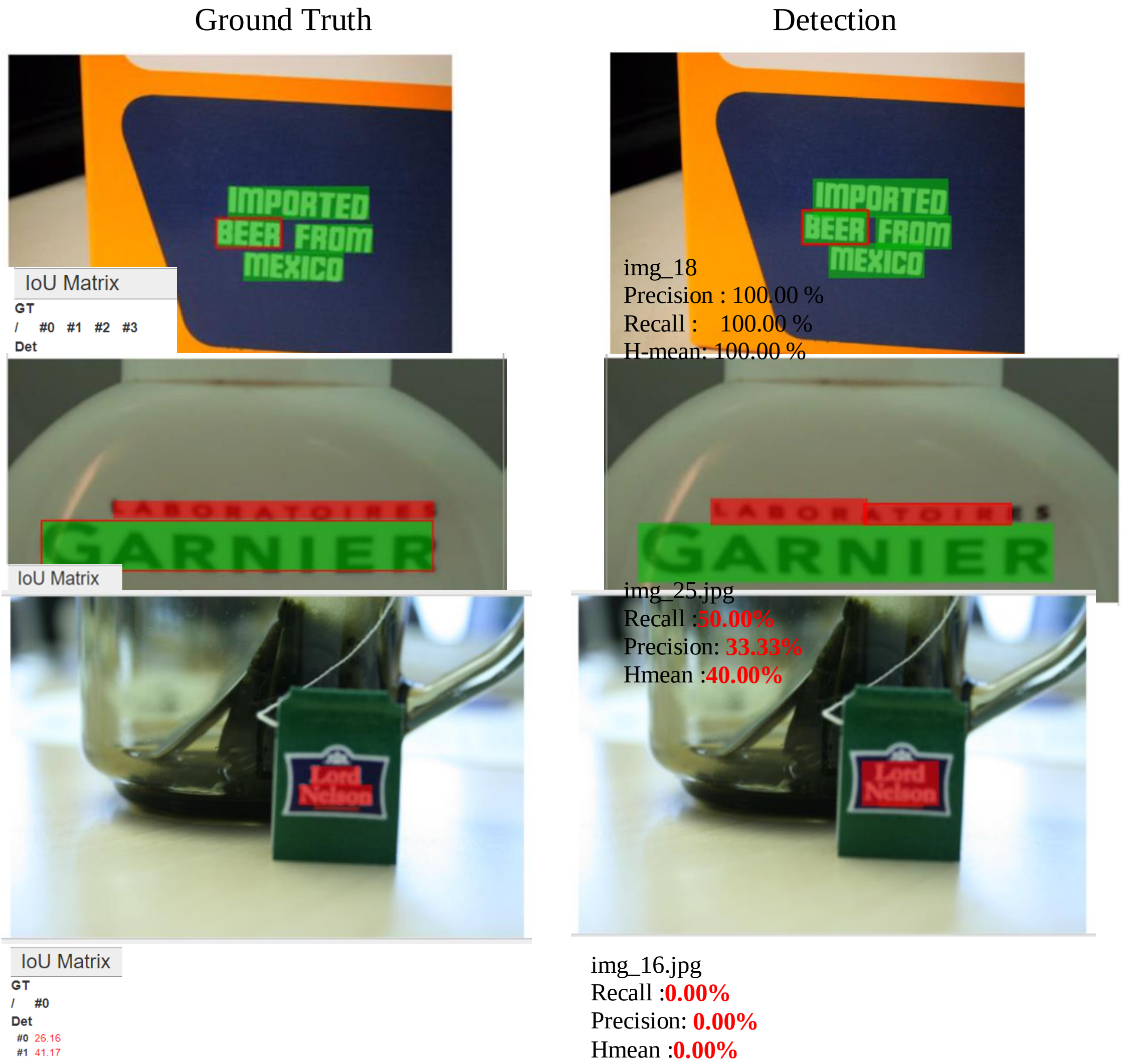

Figure 2. Sample images showing good and poor detection performance of CRAFT. Each image on the left shows ground truth while the images on the right show the results of running CRAFT.

While img_25 (middle) and img 16 (bottom) are ostensibly regarded blurry photos, img 18 (top) appears to be a clear image. CRAFT yielded a good text detection in clear images as evident by the precision, recall, and h-mean of img_18 which garnered 100%. In contrary, as seen in img_25 and img_16, text detection performance of CRAFT on blurry images produced low precision, recall, and h-mean scores. It can be inferred that the weakness of CRAFT is handling blurry instances; thus improving the blurry images can further improve CRAFT's detection performance.

### 3.2. Applying deblurring technique to *All* Images

As mentioned in the previous section, CRAFT appears to have difficulty identifying text in blurry images hence a quick workaround was made in order to attempt removing all blur in the dataset, by running a deblurring technique on *all* images in the dataset through the use of blind deconvolution.

Blind deconvolution is a technique for recovering a scene from a blurred image using an unrecognized or poorly understood point spread function (PSF). The PSF describes how much a point of light is spread out (blurred) by an optical system. The likelihood that the output recovered image is an instance of the input blurry image when convolved with a particular PSF is maximized by blind deconvolution. Every time, PSF reconstruction starts with a consistent array (array of ones) and a pair of parameter values. The number of pixels applied in each dimension (x, y) during restoration will depend on the combination of parameter values.

The first approach that has been suggested deconvolves every image by using a particular PSF. The (x, y) PSFs that were investigated in this study to enhance the text detection performance of CRAFT were those whose individual x and y values are drawn from the set of 1, 2, 3,... ,7}. However, only those with x and y below 3 showed positive results. The optical axis, which is parallel to the image's vertical axis, is represented by these paired PSF values. Applying constraints such as the result cannot be negative, or that the PSF must be symmetrical, aids selection of useful PSF models.

Figure 3 presents the original image (a) and a selected set of images (b) to (f) that were restored using blind deconvolution and the corresponding PSFs. A deblurred image first undergoes blind deconvolution, which is followed by standard text detection with CRAFT. Figure 4 further demonstrates how analyzing the reconstructed PSFs may contribute in determining the appropriate PSF values for the image.

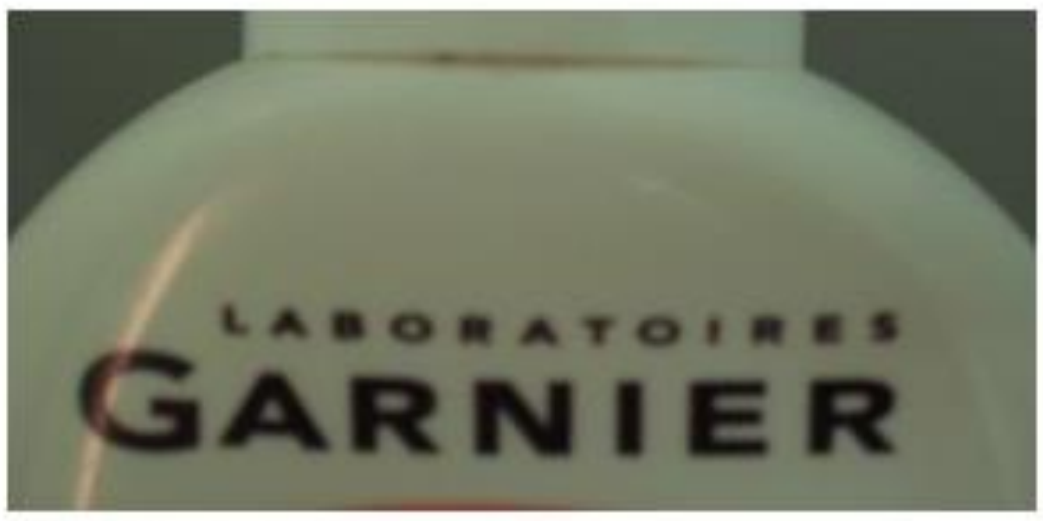

(a) Original Image

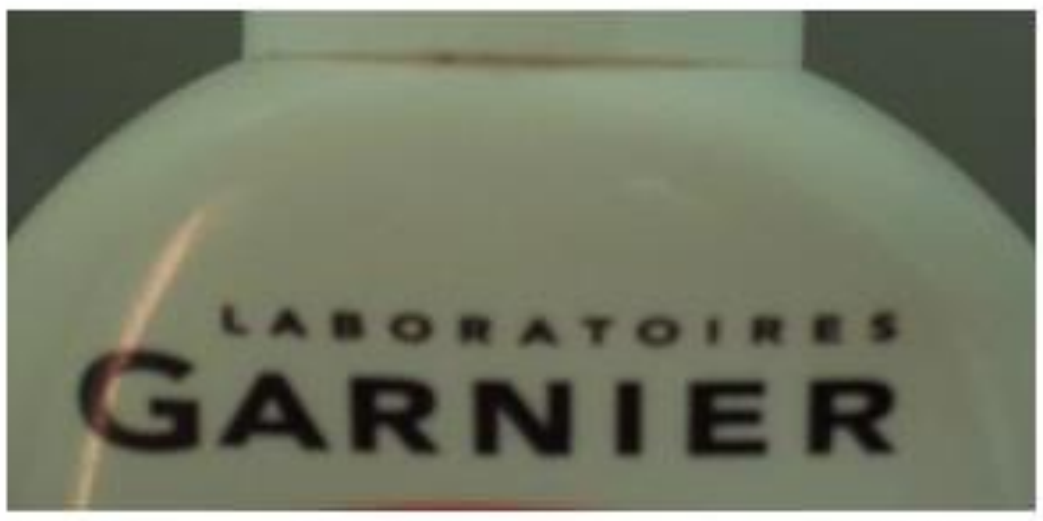

(b) Deconvolved Image using (1,1) PSF

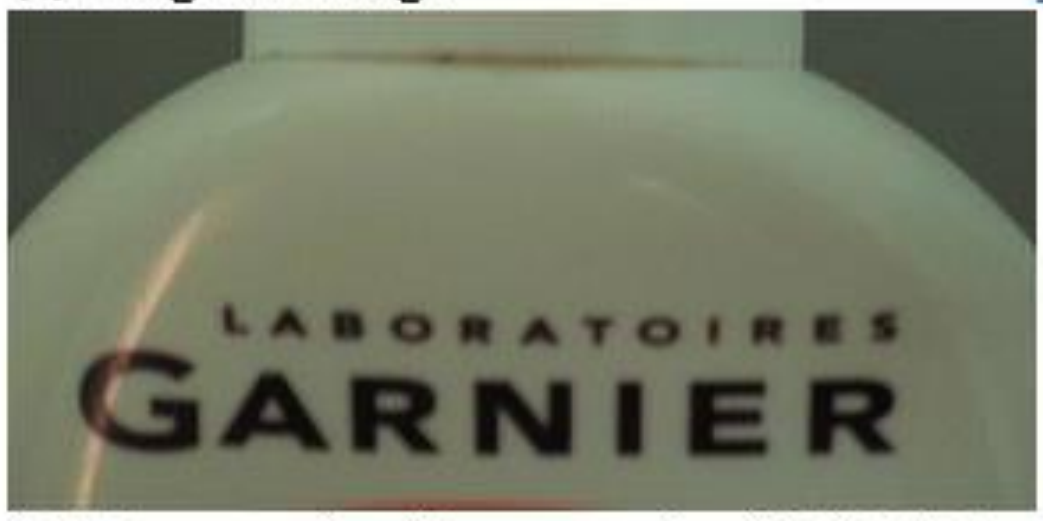

(c) Deconvolved Image using (1,2) PSF

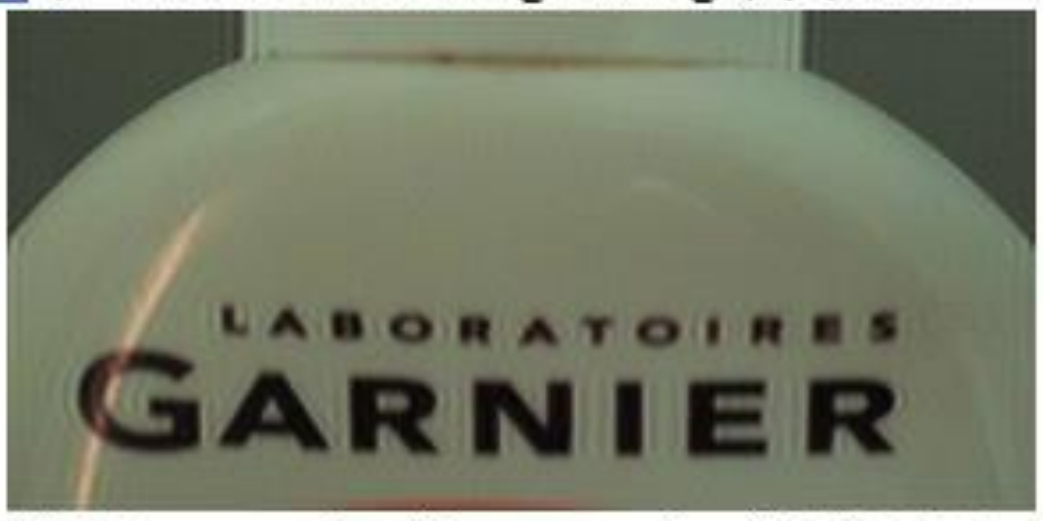

(d) Deconvolved Image using (1,3) PSF

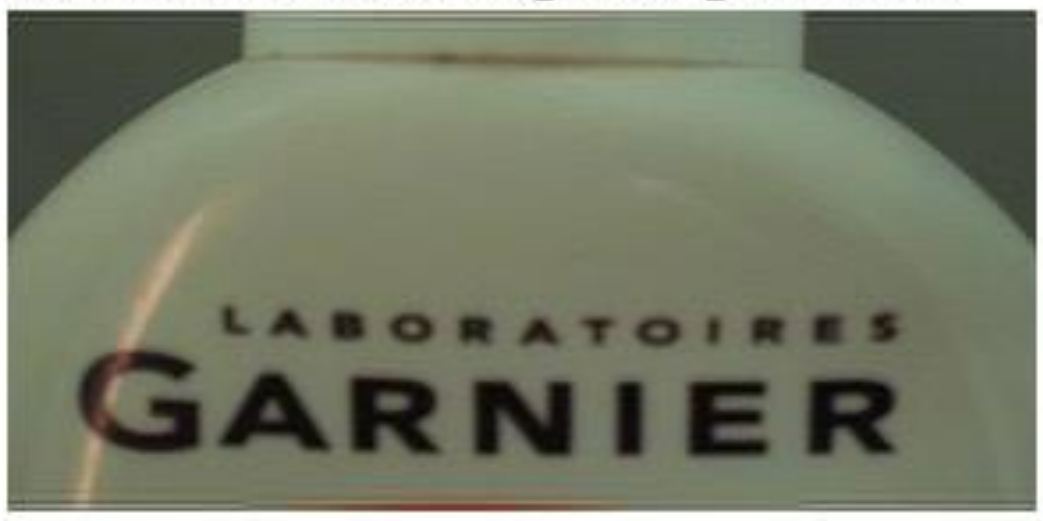

(e) Deconvolved Image using (2,1) PSF

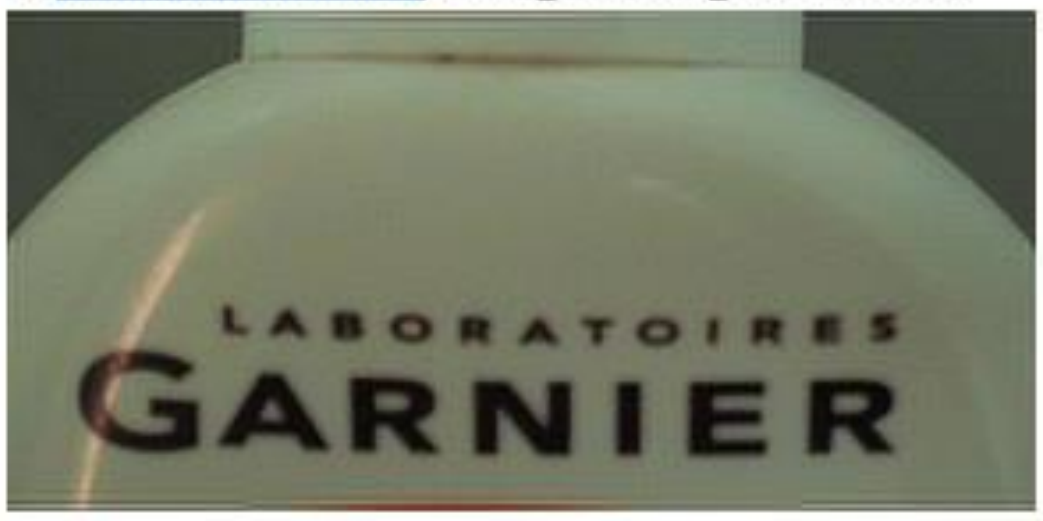

(f) Deconvolved Image using (2,2) PSF

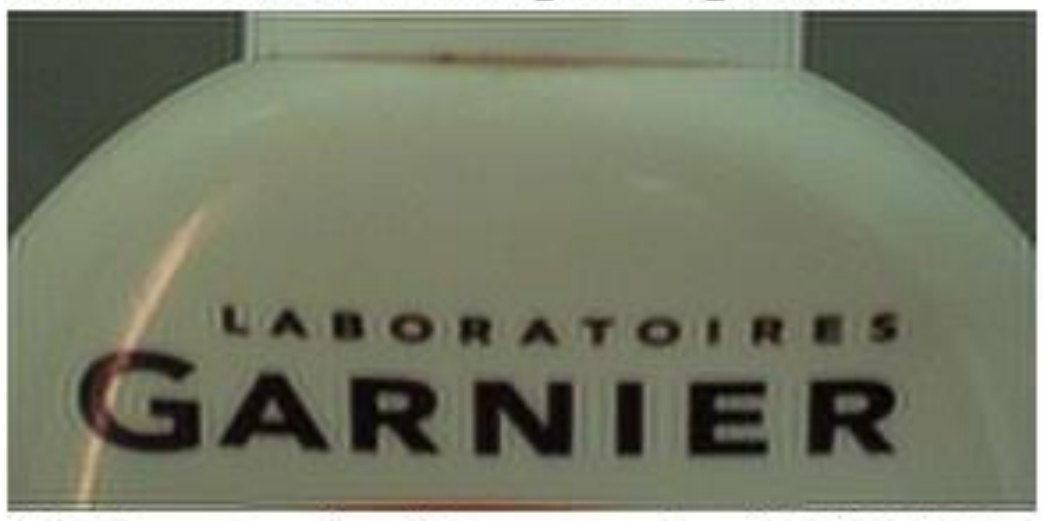

(g) Deconvolved Image using (2,3) PSF

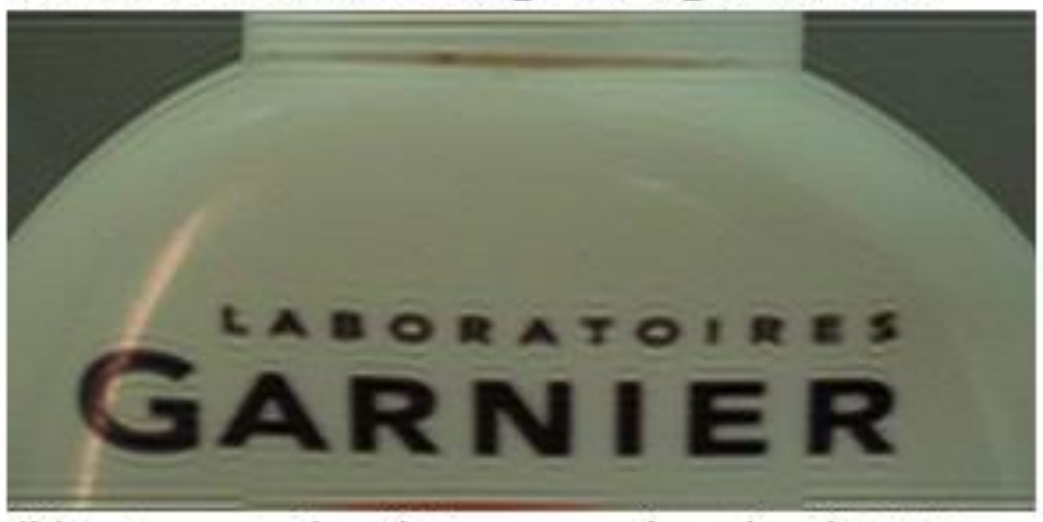

(h) Deconvolved Image using (3,1) PSF

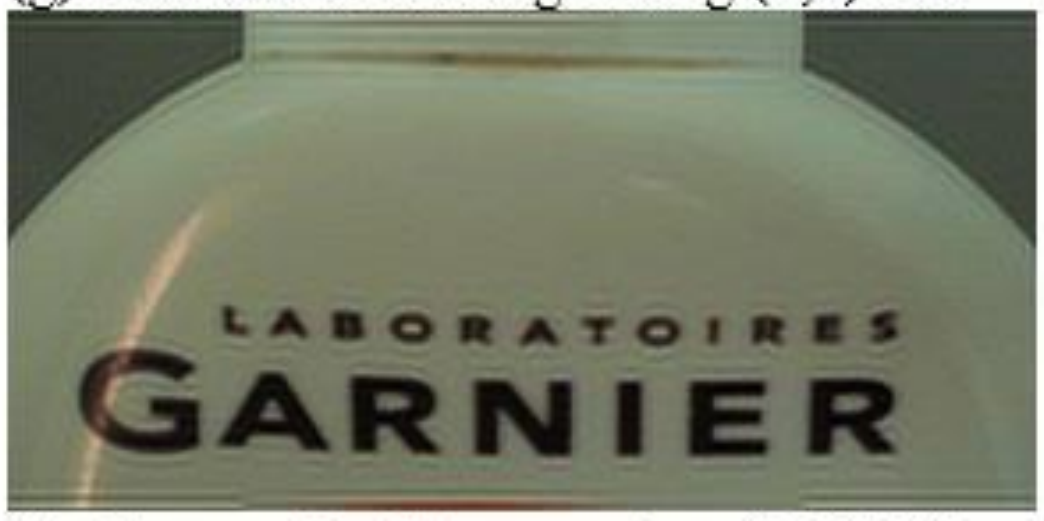

(i) Deconvolved Image using (3,2) PSF

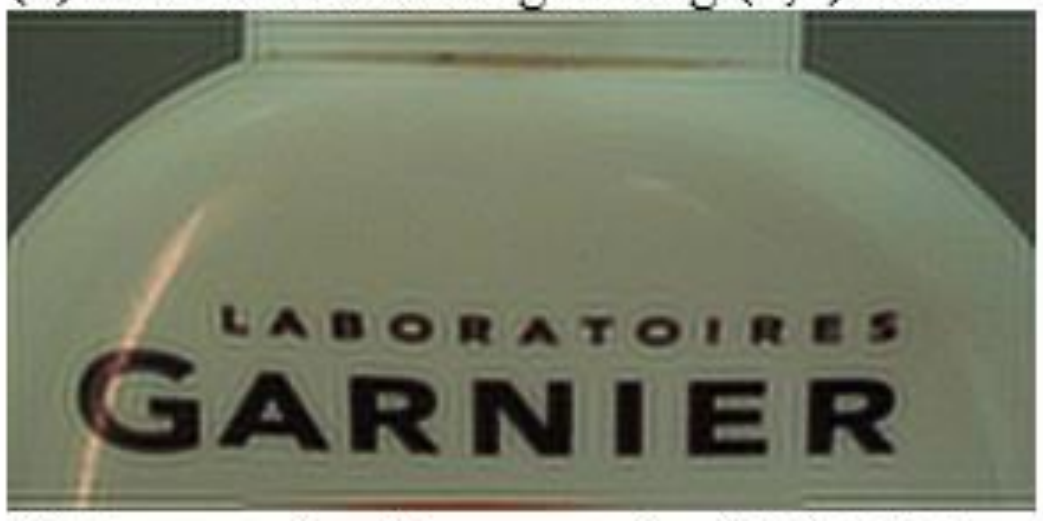

(j) Deconvolved Image using (3,3) PSF

Figure 3. Sample image which underwent blind deconvolution using different PSFs as indicated

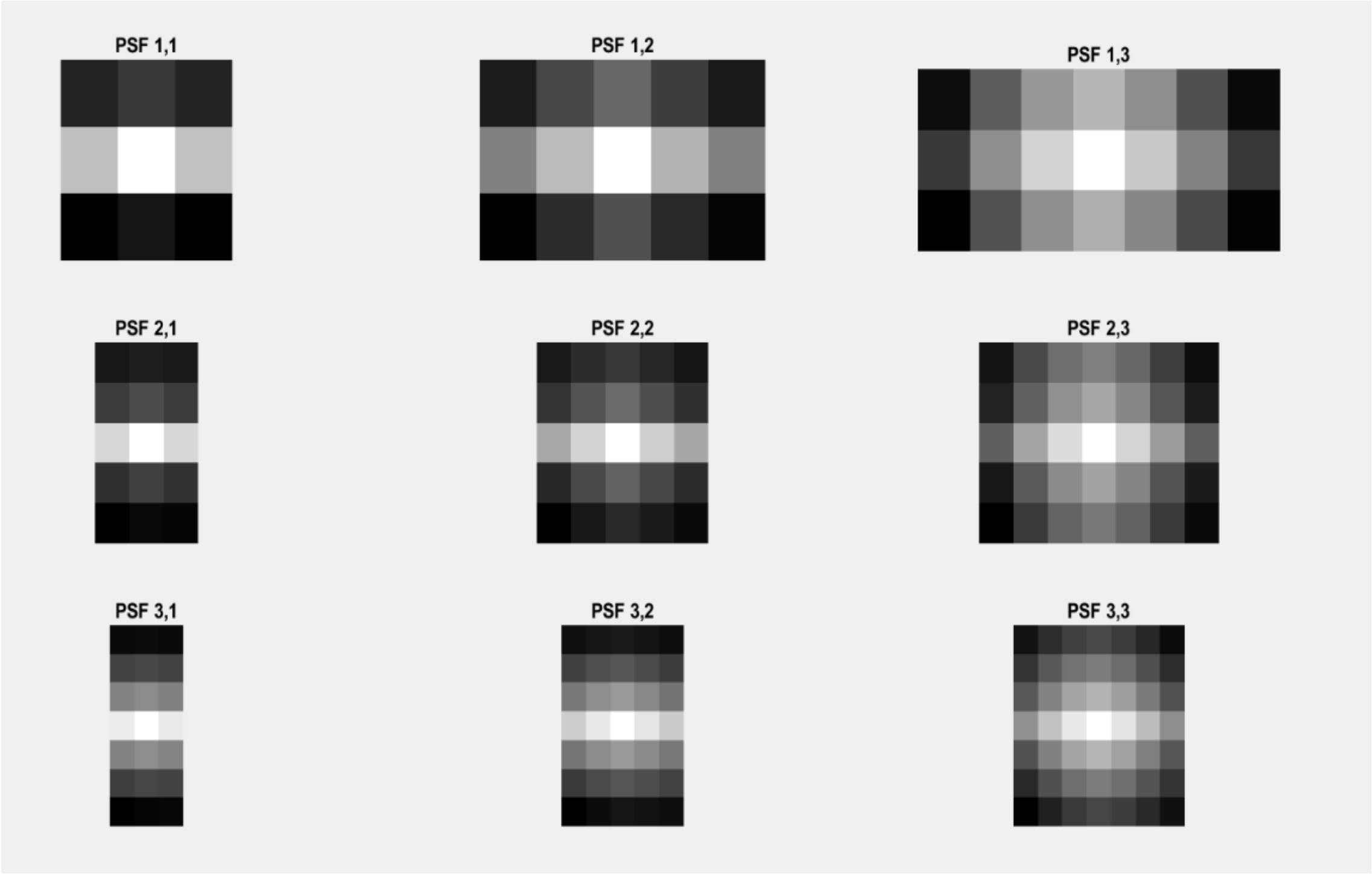

Figure 4. Reconstructed PSFs of the sample image using different PSFs as indicated

Furthermore, the results of the initial experiment, which involved blind deconvolution being applied to *all* image in the datasets, are shown in Table 2. As seen in the table, different PSFs were tried. Each entry in Table 2 shows performance that is superior to the standard CRAFT. This verifies that handling image blur is indeed a weakness in the CRAFT algorithm which was observed earlier.

Table 2. Performance of CRAFT when applying Blind Deconvolution to *All* of the Images

| **Blind Deconvolution PSF** | **Precision** | **Recall** | **H-Mean** |
|---|---|---|---|
| (1,1) | 94.42% | 92.30% | 93.35% |
| (1,2) | 94.32% | 92.44% | 93.37% |
| (1,3) | 93.78% | 91.55% | 92.65% |
| (2,1) | 93.99% | 92.09% | 93.03% |
| (2,2) | 93.62% | 91.66% | 92.63% |
| (2,3) | 93.62% | 91.66% | 92.63% |
| (3,1) | 93.28% | 91.23% | 92.24% |
| (3,2) | 93.28% | 91.23% | 92.24% |
| (3,3) | 93.62% | 91.66% | 92.63% |

The best results of using Blind Deconvolution as a pre-processing step for CRAFT was gathered when using a PSF of (1,2). The proposed CRAFT variant obtained an IoU h-mean of 93.37% under this configuration, which is a marked improvement above the 91.42% IoU h-mean of the original CRAFT. The performance is even better than the current second-ranked 93.11% from TextFuseNet. However, this is still a bit inferior to that of SenseTime, whose IoU h-mean stands at 93.62%.

An updated ranking of the top performing algorithms would show an impressive ranking of this proposed algorithm (see Table 3).

Table 3. Comparison of the State-of-the-art algorithms ranked by IoU h-mean

| Method | Precision | Recall | H-Mean |
|---|---|---|---|
| SenseTime (2016) | 91.87% | 95.45% | 93.62% |
| **BD-CRAFT v1 (2021)** | **94.32%** | **92.44%** | **93.37%** |
| TextFuseNet (2020) | 90.78% | 95.58% | 93.11% |
| TencentAILab (2017) | 94.79% | 91.37% | 93.05% |
| VARCO (2020) | 89.86% | 93.63% | 91.71% |
| HIT (2020) | 89.22% | 93.85% | 91.48% |
| CRAFT (2018) | 89.04% | 93.93% | 91.42% |

Other deblurring methods for the pre-processing step could have been explored on in order to enhance the detection performance of the CRAFT algorithm. However, as observed earlier, the CRAFT algorithm actually performs generally very well on non-blurry images. In fact, there are a few instances when the application of Blind Deconvolution to an image decreased the performance of CRAFT, as shown in Figure 5.

Figure 5. Sample image wherein CRAFT's performance decreases after applying Blind Deconvolution.

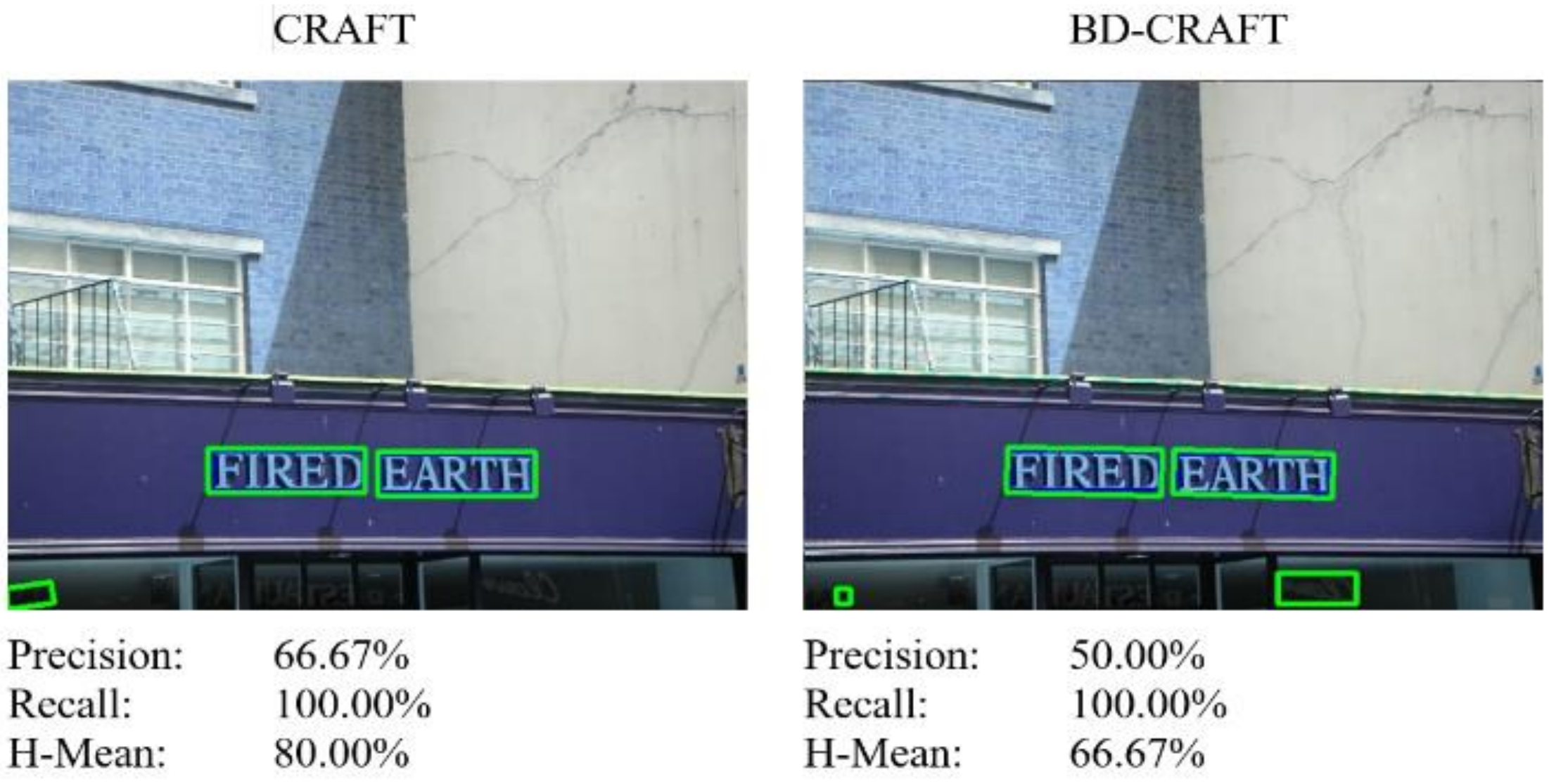

## 3.3. Separating Blurry from Non-blurry Images

The result of the previous experiment shows that, while there is significant improvement to CRAFT when all images undergo Blind Deconvolution as a pre-processing step, the proposed technique is still unable to outperform the state-of-the-art algorithm SenseTime. With these, a a technique to classify images as either blurry or non-blurry was explored on so that the deblurring technique will only be applied to images automatically detected as blurry. To achieve this, the Laplacian operator was used. It is a differential operator given by the divergence of the gradient of a scalar function on Euclidean space. Applied to images, the Laplacian enables highlighting regions of an image with rapid intensity changes. It is thus often used for edge detection.

The average variance of the Laplacian can be used to describe the blurriness of an image as a single point value. The higher the number, the sharper the edges in the image are. Therefore, this value is simply calculated for the input image and compared against a predefined threshold in order to automatically identify an image as either blurry or non-blurry.

The aforementioned threshold must be properly set as a key parameter. Applying an improper threshold or using excessive values can result in misclassification of images. Specifically, applying small thresholds may mark images as blurry when they are not while applying too large values may also mark images as non-blurry when in fact they are.

Several thresholds, including 30, 50, 80, 100, 150, and 200, were experimented on with in order to determine the most appropriate threshold for the proposed technique. The best result was obtained using threshold 100 among these. his is in consonance with the result of the published work of Pech-Pacheco [31] that a threshold 100 is a good threshold for blurry and non-blurry classification.

The reason this method works is due to the definition of the Laplacian operator itself, which is used to measure the second derivative of an image. The Laplacian, like the Sobel and Scharr operators, highlights regions of an image with rapid intensity changes. Furthermore, just like the mentioned operators, the Laplacian is often used for edge detection. The image has a large variation and a wide range of responses, both edge-like and non-edge-like, typical of a typical, in-focus image. However, if the variance is very low, there is very little spread in the responses, indicating that the image has very few edges. It is apparent that the fewer edges in an image, the blurrier it is.

After evaluating the input scene image, if the focus measure is greater than the specified threshold of 100, the image is deemed to be non-blurry, and CRAFT scene detection is then applied instantly. If not, a further pre-processing step (blind deconvolution) will be applied to the scene image.

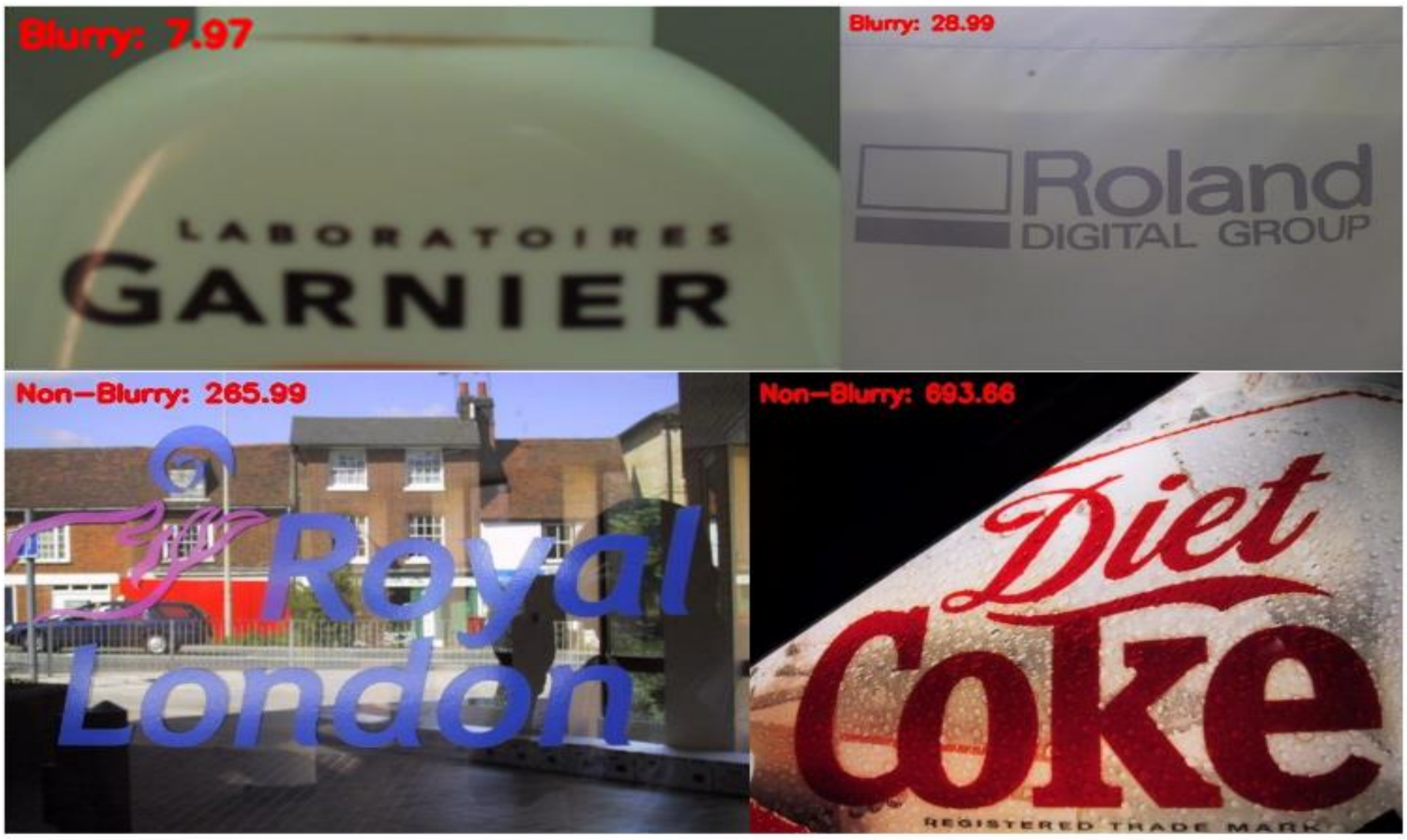

Figure 6. Example images from the ICDAR 2013 dataset, with their focused measures indicated

From the 233 images in the ICDAR 2013 dataset, exactly 63 were classified as blurry and 170 as non-blurry when the threshold 100 was used. Figure 6 shows some images which are classified as blurry and non-blurry. Refer to Figure 7 for some example images. The first two images (top) are considered blurry, with measures of 7.97 and 28.99. The last two images (bottom) are considered non-blurry, with measures of 265.99 and 693.66.

### 3.4. The Impact of Blind Deconvolution to the Identified Blurry Images

Evaluation results of the 63 identified blurry images with their respective threshold of blurriness are shown in Appendix A. The precision, recall and h-mean of the said images are also presented and were compared to the previous result of the original CRAFT.

It can be seen in Table 4 that, when blind deconvolution (BD-CRAFT) was applied to the (automatically) identified blurry images, 13 of the images had increased performance (h-mean), while 11 of the images had improved recall and precision compared to the results when using the original CRAFT algorithm. Furthermore, it has 27.51% average h-mean improvement while earning 29.79% and 23.61% average increase to its precision and recall respectively. It can be noticed that its precision has the gained highest increase as reflected by the evaluation results. On the other hand, 7 images yielded lower h-mean results when using BD-CRAFT compared to the performance of the said images when the standard CRAFT is used. It yielded an average decreased h-mean performance of 20.10% while 26.29% and 7.14% to its precision and recall. The number of images with increased performance as well as the average increase in its h-mean results show that, overall, CRAFT is improved when Blind Deconvolution is performed, and we refer to this improved version as BD-CRAFT.

Table 4. Impact of Using Blind Deconvolution (BD-CRAFT) vs CRAFT to Some Automatically Identified Blurry Images

| Blurry Images | Blurriness | CRAFT | | | BD-CRAFT | | |
|---|---|---|---|---|---|---|---|
| **Img_no** | **Threshold** | **Precision** | **Recall** | **H-Mean** | **Precision** | **Recall** | **H-Mean** |
| 4 | 16.4762852 | 80.00% | 57.14% | 66.67% | **100.00%** | **85.71%** | **92.31%** |
| 16 | 16.1522402 | 0.00% | 0.00% | 0.00% | **100.00%** | **50.00%** | **66.67%** |
| 23 | 15.9194037 | 100.00% | 100.00% | 100.00% | **94.12%** | **100.00%** | **96.87%** |
| 25 | 7.97007126 | 33.33% | 50.00% | 40.00% | **100.00%** | **100.00%** | **100.00%** |
| 26 | 35.1926793 | 90.91% | 76.92% | 83.33% | **100.00%** | **100.00%** | **100.00%** |
| 29 | 50.7909314 | 100.00% | 60.00% | 75.00% | **75.00%** | **60.00%** | **66.67%** |
| 34 | 16.1037522 | 100.00% | 100.00% | 100.00% | **45.45%** | **100.00%** | **62.50%** |
| 39 | 15.4544795 | 88.89% | 80.00% | 84.21% | **83.33%** | **100.00%** | **90.91%** |
| 45 | 21.6460108 | 76.47% | 86.67% | 81.25% | **100.00%** | **93.33%** | **96.55%** |
| 49 | 56.3667426 | 100.00% | 66.67% | 80.00% | **100.00%** | **83.33%** | **90.90%** |
| 54 | 39.2921302 | 71.43% | 83.33% | 76.92% | **100.00%** | **100.00%** | **100.00%** |
| 69 | 28.2970963 | 100.00% | 100.00% | 100.00% | **50.00%** | **100.00%** | **66.67%** |
| 84 | 22.8809663 | 100.00% | 83.33% | 90.91% | **71.43%** | **83.33%** | **76.92%** |
| 87 | 49.7199117 | 100.00% | 100.00% | 100.00% | **80.00%** | **100.00%** | **88.89%** |
| 93 | 13.5899878 | 50.00% | 100.00% | 66.67% | **100.00%** | **100.00%** | **100.00%** |
| 125 | 29.4063831 | 83.33% | 71.43% | 76.92% | **100.00%** | **100.00%** | **100.00%** |
| 183 | 81.9879789 | 100.00% | 100.00% | 100.00% | **100.00%** | **50.00%** | **66.67%** |
| 184 | 26.0136719 | 80.00% | 100.00% | 88.89% | **100.00%** | **100.00%** | **100.00%** |
| 231 | 60.5148326 | 50.00% | 33.33% | 40.00% | **83.33%** | **100.00%** | **90.91%** |
| 232 | 17.4316837 | 75.00% | 100.00% | 85.71% | **100.00%** | **100.00%** | **100.00%** |

Meanwhile, Table 5 shows the performance of some identified blurry images using BD-CRAFT when compared to SenseTime. The table shows that when BD-CRAFT was applied, 11 images performed better in terms of h-mean than Sensetime, whereas 8 and 9 images performed better in terms of precision and recall. It has earned an average increase of 21.83% as to its h-mean performance while garnering 21.51% and 18.47% to its precision and recall. However, there are also 9 images which gained lower h-mean results when BD-CRAFT is applied which has a reduced performance of 17.69% to its h-mean result and 13.39% and 12.71% to its precision and recall respectively. The images which have improved results when using BD-CRAFT outnumbered the images which had lower results than SenseTime. This partly explains why BD-CRAFT is better than SenseTime overall. The evaluation results of the 63 automatically identified blurry images using BD-CRAFT and SenseTime is presented in Appendix B.

Table 5. Impact of Using Blind Deconvolution (BD-CRAFT) vs Sensetime to Some Automatically Identified Blurry Images

| **Blurry Images** | **Blurriness** | **BD-CRAFT** | | | **SenseTime** | | |
|---|---|---|---|---|---|---|---|
| **Img_no.** | **Threshold** | **Precision** | **Recall** | **H-Mean** | **Precision** | **Recall** | **H-Mean** |
| 4 | 16.4762852 | **100.00%** | **85.71%** | **92.31%** | 71.43% | 71.43% | 71.43% |
| 16 | 16.1522402 | **100.00%** | **50.00%** | **66.67%** | 100.00% | 100.00% | 100.00% |
| 23 | 15.9194037 | **94.12%** | **100.00%** | **96.87%** | 100.00% | 100.00% | 100.00% |
| 26 | 35.1926793 | **100.00%** | **100.00%** | **100.00%** | 100.00% | 84.62% | 91.67% |
| 29 | 50.7909314 | **75.00%** | **60.00%** | **66.67%** | 80.00% | 80.00% | 80.00% |
| 34 | 16.1037522 | **45.45%** | **100.00%** | **62.50%** | 100.00% | 100.00% | 100.00% |
| 38 | 14.3117793 | **100.00%** | **100.00%** | **100.00%** | 50.00% | 100.00% | 66.67% |
| 39 | 15.4544795 | **83.33%** | **100.00%** | **90.91%** | 66.67% | 80.00% | 72.73% |
| 45 | 21.6460108 | **100.00%** | **93.33%** | **96.55%** | 100.00% | 100.00% | 100.00% |
| 49 | 56.3667426 | **100.00%** | **83.33%** | **90.90%** | 75.00% | 75.00% | 75.00% |
| 59 | 97.4825748 | **100.00%** | **100.00%** | **100.00%** | 75.00% | 75.00% | 75.00% |
| 69 | 28.2970963 | **50.00%** | **100.00%** | **66.67%** | 100.00% | 100.00% | 100.00% |
| 77 | 88.4263075 | **100.00%** | **100.00%** | **100.00%** | 50.00% | 50.00% | 50.00% |
| 84 | 22.8809663 | **71.43%** | **83.33%** | **76.92%** | 80.00% | 66.67% | 72.73% |
| 85 | 68.5711141 | **100.00%** | **100.00%** | **100.00%** | 50.00% | 100.00% | 66.67% |
| 125 | 29.4063831 | **100.00%** | **100.00%** | **100.00%** | 100.00% | 71.43% | 83.33% |
| 183 | 81.9879789 | **100.00%** | **50.00%** | **66.67%** | 75.00% | 75.00% | 75.00% |
| 227 | 44.3401323 | **100.00%** | **100.00%** | **100.00%** | 100.00% | 75.00% | 85.71% |
| 231 | 60.5148326 | **83.33%** | **100.00%** | **90.91%** | 100.00% | 100.00% | 100.00% |

Furthermore, the experimental findings on Table 6 demonstrate that the performance of CRAFT improves in specific PSF values such as (1,1), (1,2), (1,3), (2,1), (3,2), and when blind deconvolution is applied only to recognized blurry images (3,3). Using the PSF, blind deconvolved images produced the best results (1,3). This achieved an h-mean of 94.47%, surpassing the 93.62% obtained by the state-of-the-art SenseTime. There are other PSF values that eventually yielded results also better than SenseTime's. This supports the theory that, in cases when the PSF is unknown, it may be able to determine it by repeatedly testing several potential PSFs and evaluating whether the image has improved.

Table 6. Performance of CRAFT when applying Blind Deconvolution (BD) to the detected blur images

| **Blind Deconvolution PSF** | **Precision** | **Recall** | **H-Mean** |
|---|---|---|---|
| **(1,1)** | **94.68%** | **93.63%** | **94.15%** |
| **(1,2)** | **94.54%** | **93.92%** | **94.24%** |
| **(1,3)** | **95.24%** | **93.72%** | **94.47%** |
| **(2,1)** | **94.48%** | **93.42%** | **93.94%** |
| **(2,2)** | **94.61%** | **93.61%** | **94.10%** |
| (2,3) | 94.00% | 92.91% | 93.45% |
| (3,1) | 92.86% | 92.39% | 92.61% |
| **(3,2)** | **94.43%** | **93.35%** | **93.88%** |
| **(3,3)** | **94.35%** | **93.10%** | **93.72%** |

Using (1,3) as the PSF for BD-CRAFT, the updated ranking of the state-of-the-art algorithms for Text Detection shows BD-CRAFT ranked top (see Table 7).

Note that BD-CRAFT outperforms our first suggested method (which employs Blind Deconvolution on all images in the dataset) on every performance metric. Its IoU h-mean is 0.85% (absolute) greater than SenseTime and 3.05% higher than the original CRAFT. The highest precision result in the table, 95.24%, which outperforms SenseTime by more than 3%, is what prompted the performance to achieve the top rating.

Table 7. Comparison of the State-of-the-art algorithms ranked by IoU h-mean

| Method | Precision | Recall | H-Mean |
|---|---|---|---|
| **Our proposed BD-CRAFT (2021)** | **95.24%** | **93.72%** | **94.47%** |
| SenseTime (2016) | 91.87% | 95.45% | 93.62% |
| **Our first proposed method (2021)** | **94.32%** | **92.44%** | **93.37%** |
| TextFuseNet (2020) | 90.78% | 95.58% | 93.11% |
| TencentAILab (2017) | 94.79% | 91.37% | 93.05% |
| VARCO (2020) | 89.86% | 93.63% | 91.71% |
| HIT (2020) | 89.22% | 93.85% | 91.48% |
| CRAFT (2018) | 89.04% | 93.93% | 91.42% |

## 3.5. The Proposed BD-CRAFT Algorithm for Scene Text Detection

Using the insights gathered from previous experiments, BD-CRAFT, an improved variant of CRAFT for scene text detection is proposed. A flowchart describing the main operations of BD-CRAFT is provided in Figure 7.

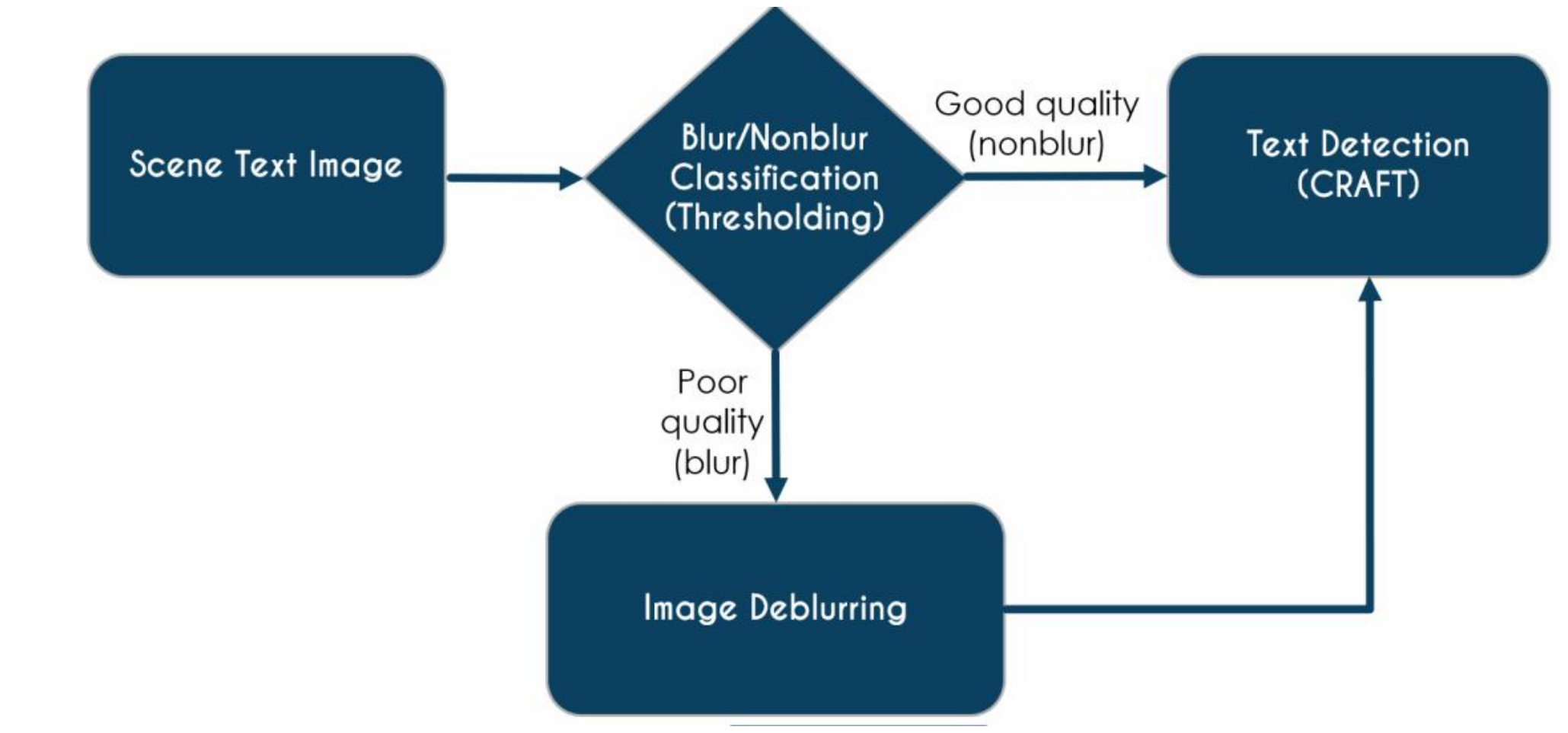

Figure 7. Flowchart of the proposed BD-CRAFT

*Blurry/Non-blurry Classification.* In this phase, the scene text image is automatically classified as either blurry or non-blurry using the Laplacian operator and a threshold of 100.

*Blind deconvolution.* The blind deconvolution is used to deblur images that have been identified as blurry. Note that those images that are classified as non-blurry skip this step. Based on the experimental results, PSF (1,3) is the ideal PSF in deblurring the images as it produce better results than the other PSFs.

*Text Detection using CRAFT.* After preprocessing the blurry images using blind deconvolution, the text detection step takes place. In detecting text areas or regions, the Character Region Awareness for Text Detection (CRAFT) was used in this study. CRAFT is a text detection method which effectively detects text areas by exploring each character and affinity between characters. Its primary objective is to locate individual characters in natural images. A deep neural network is trained to predict character regions as well as character affinity to cover a variety of text shapes in a bottom-up manner. The key strength of CRAFT as a novel text detector is that it can detect individual characters even if the character-level annotation is unknown.

With this, blurry and non-blurry image classification is considered imploring Laplacian of variance with 100 threshold. The detected non-blurry images will then undergo scene detection using CRAFT while the blurry scene image will further undergo blind deconvolution before imploring CRAFT.

## 4. CONCLUSION AND RECOMMENDATION

A variant of the CRAFT algorithm for scene text detection called BD-CRAFT was developed in this study. This variant involves the pre-processing steps of automatic image classification using Laplacian operator, with threshold 100 followed by the image deblurring using Blind Deconvolution of detected blurry images using PSF (1,3). By combining said pre-processing steps with CRAFT, the BD-CRAFT improves the performance of CRAFT sufficiently to the point of besting even the SenseTime algorithm which is the current top-ranking state-of-the-art algorithm for scene text detection. In ICDAR 2013, BD-CRAFT garnered an h-mean of 94.47%, which is better than those of CRAFT (91.42%) and SenseTime (93.62%). As possibilities for further studies, it would be interesting to investigate which state-of-the-art algorithms for scene text detection would benefit also from the discussed input pre-processing techniques. It may also be good to investigate other pre-processing techniques that can be incorporated in text detection algorithms in general.

### ACKNOWLEDGEMENTS

The completion of this undertaking could not have been possible without the participation and assistance of many individuals whose names may not all be enumerated, their contributions are sincerely appreciated and gratefully acknowledged. Through the guidance, blessings and graces of the Lord Almighty, the preparation of this undertaking was made possible. It is with deep appreciation that we offer this piece of work to CHED, Ateneo de Manila University and DMMMSU for their support towards the fulfillment of this scholarly endeavor. It is the kindness of these acknowledged that this research sees the light of the day. We submit this work with great humility and utmost regard.

## AUTHORS

**Clarisa V. Albarillo** is an Assistant Professor III of the Don Mariano Marcos Memorial State University and currently taking up PhD in Computer Science at the Ateneo de Manila University. She is a research enthusiast in the field of artificial intelligence, computer vision, image processing, and social research.

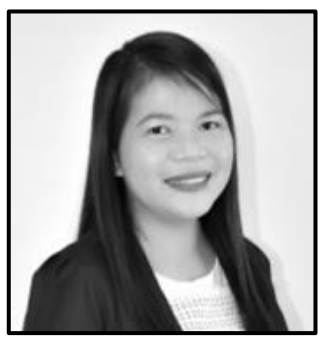

**Proceso L. Fernandez Jr**. is a Professor of Computer Science at the Ateneo de Manila University. His main research interests include those in the areas of Computer Vision, Machine Learning, and Algorithms and Complexity Theory. He currently also serves as the Director of the Ateneo Intellectual Property Office of said University.

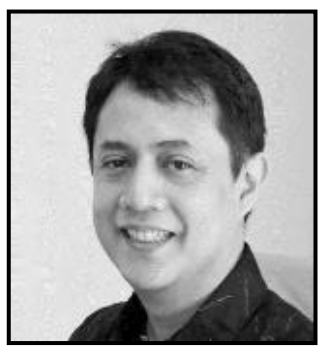

Appendix A. Evaluation Performance of the 63 Blurry Images using BD-CRAFT vs CRAFT

| **Blurry Images** | **Blurriness** | **CRAFT** | | | **BD-CRAFT** | | |
|---|---|---|---|---|---|---|---|
| **Img_no.** | **Threshold** | **Precision** | **Recall** | **H-Mean** | **Precision** | **Recall** | **H-Mean** |
| 1 | 12.8405571 | 100.00% | 100.00% | 100.00% | 100.00% | 100.00% | 100.00% |
| 4 | 16.4762852 | 80.00% | 57.14% | 66.67% | **100.00%** | **85.71%** | **92.31%** |
| 10 | 23.4311781 | 100.00% | 100.00% | 100.00% | 100.00% | 100.00% | 100.00% |
| 12 | 36.653517 | 100.00% | 100.00% | 100.00% | 100.00% | 100.00% | 100.00% |
| 15 | 28.9903253 | 100.00% | 100.00% | 100.00% | 100.00% | 100.00% | 100.00% |
| 16 | 16.1522402 | 0.00% | 0.00% | 0.00% | **100.00%** | **50.00%** | **66.67%** |
| 20 | 57.4975666 | 100.00% | 100.00% | 100.00% | 100.00% | 100.00% | 100.00% |
| 21 | 81.7039225 | 100.00% | 100.00% | 100.00% | 100.00% | 100.00% | 100.00% |
| 23 | 15.9194037 | 100.00% | 100.00% | 100.00% | **94.12%** | **100.00%** | **96.87%** |
| 25 | 7.97007126 | 33.33% | 50.00% | 40.00% | **100.00%** | **100.00%** | **100.00%** |
| 26 | 35.1926793 | 90.91% | 76.92% | 83.33% | **100.00%** | **100.00%** | **100.00%** |
| 29 | 50.7909314 | 100.00% | 60.00% | 75.00% | **75.00%** | **60.00%** | **66.67%** |
| 34 | 16.1037522 | 100.00% | 100.00% | 100.00% | **45.45%** | **100.00%** | **62.50%** |
| 38 | 14.3117793 | 100.00% | 100.00% | 100.00% | **100.00%** | **100.00%** | **100.00%** |
| 39 | 15.4544795 | 88.89% | 80.00% | 84.21% | **83.33%** | **100.00%** | **90.91%** |
| 44 | 27.4993814 | 100.00% | 100.00% | 100.00% | 100.00% | 100.00% | 100.00% |
| 45 | 21.6460108 | 76.47% | 86.67% | 81.25% | **100.00%** | **93.33%** | **96.55%** |
| 48 | 17.6449657 | 100.00% | 100.00% | 100.00% | 100.00% | 100.00% | 100.00% |
| 49 | 56.3667426 | 100.00% | 66.67% | 80.00% | **100.00%** | **83.33%** | **90.90%** |
| 52 | 99.1999698 | 100.00% | 100.00% | 100.00% | 100.00% | 100.00% | 100.00% |
| 54 | 39.2921302 | 71.43% | 83.33% | 76.92% | **100.00%** | **100.00%** | **100.00%** |
| 55 | 83.6990885 | 100.00% | 100.00% | 100.00% | 100.00% | 100.00% | 100.00% |
| 58 | 41.8884998 | 100.00% | 77.78% | 87.50% | 100.00% | 77.78% | 87.50% |
| 59 | 97.4825748 | 100.00% | 100.00% | 100.00% | 100.00% | 100.00% | 100.00% |
| 61 | 61.5662565 | 100.00% | 100.00% | 100.00% | 100.00% | 100.00% | 100.00% |
| 62 | 64.4650431 | 100.00% | 100.00% | 100.00% | 100.00% | 100.00% | 100.00% |
| 63 | 66.767582 | 100.00% | 100.00% | 100.00% | 100.00% | 100.00% | 100.00% |
| 64 | 18.0964371 | 100.00% | 100.00% | 100.00% | 100.00% | 100.00% | 100.00% |
| 65 | 12.5359277 | 100.00% | 100.00% | 100.00% | 100.00% | 100.00% | 100.00% |
| 69 | 28.2970963 | 100.00% | 100.00% | 100.00% | **50.00%** | **100.00%** | **66.67%** |
| 73 | 78.597045 | 100.00% | 100.00% | 100.00% | 100.00% | 100.00% | 100.00% |
| 75 | 62.6589596 | 100.00% | 100.00% | 100.00% | 100.00% | 100.00% | 100.00% |
| 77 | 88.4263075 | 100.00% | 100.00% | 100.00% | 100.00% | 100.00% | 100.00% |
| 82 | 56.0501649 | 100.00% | 100.00% | 100.00% | 100.00% | 100.00% | 100.00% |
| 84 | 22.8809663 | 100.00% | 83.33% | 90.91% | **71.43%** | **83.33%** | **76.92%** |
| 85 | 68.5711141 | 100.00% | 100.00% | 100.00% | 100.00% | 100.00% | 100.00% |
| 87 | 49.7199117 | 100.00% | 100.00% | 100.00% | **80.00%** | **100.00%** | **88.89%** |
| 88 | 81.8170171 | 100.00% | 100.00% | 100.00% | 100.00% | 100.00% | 100.00% |
| 89 | 14.206154 | 100.00% | 100.00% | 100.00% | 100.00% | 100.00% | 100.00% |
| 93 | 13.5899878 | 50.00% | 100.00% | 66.67% | **100.00%** | **100.00%** | **100.00%** |
| 95 | 57.426799 | 100.00% | 100.00% | 100.00% | 100.00% | 100.00% | 100.00% |
| 96 | 11.8740355 | 100.00% | 100.00% | 100.00% | 100.00% | 100.00% | 100.00% |
| 98 | 61.3983121 | 100.00% | 100.00% | 100.00% | 100.00% | 100.00% | 100.00% |
| 123 | 54.5686799 | 100.00% | 100.00% | 100.00% | 100.00% | 100.00% | 100.00% |
| 125 | 29.4063831 | 83.33% | 71.43% | 76.92% | **100.00%** | **100.00%** | **100.00%** |
| 131 | 65.3736027 | 100.00% | 100.00% | 100.00% | 100.00% | 100.00% | 100.00% |
| 134 | 45.5904438 | 100.00% | 100.00% | 100.00% | 100.00% | 100.00% | 100.00% |
| 138 | 72.2330041 | 100.00% | 100.00% | 100.00% | 100.00% | 100.00% | 100.00% |
| 142 | 73.576066 | 100.00% | 100.00% | 100.00% | 100.00% | 100.00% | 100.00% |
| 143 | 24.0986572 | 100.00% | 100.00% | 100.00% | 100.00% | 100.00% | 100.00% |
| 180 | 10.3440649 | 100.00% | 100.00% | 100.00% | 100.00% | 100.00% | 100.00% |
| 181 | 84.1780428 | 100.00% | 100.00% | 100.00% | 100.00% | 100.00% | 100.00% |
| 183 | 81.9879789 | 100.00% | 100.00% | 100.00% | **100.00%** | **50.00%** | **66.67%** |
| 184 | 26.0136719 | 80.00% | 100.00% | 88.89% | **100.00%** | **100.00%** | **100.00%** |
| 186 | 35.5071598 | 100.00% | 100.00% | 100.00% | 100.00% | 100.00% | 100.00% |
| 211 | 22.1367073 | 100.00% | 100.00% | 100.00% | 100.00% | 100.00% | 100.00% |
| 217 | 95.5176921 | 100.00% | 100.00% | 100.00% | 100.00% | 100.00% | 100.00% |
| 222 | 65.4059773 | 100.00% | 100.00% | 100.00% | 100.00% | 100.00% | 100.00% |
| 226 | 43.1260097 | 100.00% | 100.00% | 100.00% | 100.00% | 100.00% | 100.00% |
| 227 | 44.3401323 | 100.00% | 100.00% | 100.00% | 100.00% | 100.00% | 100.00% |
| 229 | 16.0993804 | 100.00% | 100.00% | 100.00% | 100.00% | 100.00% | 100.00% |
| 231 | 60.5148326 | 50.00% | 33.33% | 40.00% | **83.33%** | **100.00%** | **90.91%** |
| 232 | 17.4316837 | 75.00% | 100.00% | 85.71% | **100.00%** | **100.00%** | **100.00%** |

Appendix B. Evaluation Performance of the 63 Blurry Images using BD-CRAFT vs SenseTime

| **Blurry Images** | **Blurriness** | **BD-CRAFT** | | | **SenseTime** | | |
|---|---|---|---|---|---|---|---|
| **Img_no.** | **Threshold** | **Precision** | **Recall** | **H-Mean** | **Precision** | **Recall** | **H-Mean** |
| 1 | 12.8405571 | 100.00% | 100.00% | 100.00% | 100.00% | 100.00% | 100.00% |
| 4 | 16.4762852 | **100.00%** | **85.71%** | **92.31%** | 71.43% | 71.43% | 71.43% |
| 10 | 23.4311781 | 100.00% | 100.00% | 100.00% | 100.00% | 100.00% | 100.00% |
| 12 | 36.653517 | 100.00% | 100.00% | 100.00% | 100.00% | 100.00% | 100.00% |
| 15 | 28.9903253 | 100.00% | 100.00% | 100.00% | 100.00% | 100.00% | 100.00% |
| 16 | 16.1522402 | **100.00%** | **50.00%** | **66.67%** | 100.00% | 100.00% | 100.00% |
| 20 | 57.4975666 | 100.00% | 100.00% | 100.00% | 100.00% | 100.00% | 100.00% |
| 21 | 81.7039225 | 100.00% | 100.00% | 100.00% | 100.00% | 100.00% | 100.00% |
| 23 | 15.9194037 | **94.12%** | **100.00%** | **96.87%** | 100.00% | 100.00% | 100.00% |
| 25 | 7.97007126 | 100.00% | 100.00% | 100.00% | 100.00% | 100.00% | 100.00% |
| 26 | 35.1926793 | **100.00%** | **100.00%** | **100.00%** | 100.00% | 84.62% | 91.67% |
| 29 | 50.7909314 | **75.00%** | **60.00%** | **66.67%** | 80.00% | 80.00% | 80.00% |
| 34 | 16.1037522 | **45.45%** | **100.00%** | **62.50%** | 100.00% | 100.00% | 100.00% |
| 38 | 14.3117793 | **100.00%** | **100.00%** | **100.00%** | 50.00% | 100.00% | 66.67% |
| 39 | 15.4544795 | **83.33%** | **100.00%** | **90.91%** | 66.67% | 80.00% | 72.73% |
| 44 | 27.4993814 | 100.00% | 100.00% | 100.00% | 100.00% | 100.00% | 100.00% |
| 45 | 21.6460108 | **100.00%** | **93.33%** | **96.55%** | 100.00% | 100.00% | 100.00% |
| 48 | 17.6449657 | 100.00% | 100.00% | 100.00% | 100.00% | 100.00% | 100.00% |
| 49 | 56.3667426 | **100.00%** | **83.33%** | **90.90%** | 75.00% | 75.00% | 75.00% |
| 52 | 99.1999698 | 100.00% | 100.00% | 100.00% | 100.00% | 100.00% | 100.00% |
| 54 | 39.2921302 | 100.00% | 100.00% | 100.00% | 100.00% | 100.00% | 100.00% |
| 55 | 83.6990885 | 100.00% | 100.00% | 100.00% | 100.00% | 100.00% | 100.00% |
| 58 | 41.8884998 | 100.00% | 77.78% | 87.50% | 100.00% | 77.78% | 87.50% |
| 59 | 97.4825748 | **100.00%** | **100.00%** | **100.00%** | 75.00% | 75.00% | 75.00% |
| 61 | 61.5662565 | 100.00% | 100.00% | 100.00% | 100.00% | 100.00% | 100.00% |
| 62 | 64.4650431 | 100.00% | 100.00% | 100.00% | 100.00% | 100.00% | 100.00% |
| 63 | 66.767582 | 100.00% | 100.00% | 100.00% | 100.00% | 100.00% | 100.00% |
| 64 | 18.0964371 | 100.00% | 100.00% | 100.00% | 100.00% | 100.00% | 100.00% |
| 65 | 12.5359277 | 100.00% | 100.00% | 100.00% | 100.00% | 100.00% | 100.00% |
| 69 | 28.2970963 | **50.00%** | **100.00%** | **66.67%** | 100.00% | 100.00% | 100.00% |
| 73 | 78.597045 | 100.00% | 100.00% | 100.00% | 100.00% | 100.00% | 100.00% |
| 75 | 62.6589596 | 100.00% | 100.00% | 100.00% | 100.00% | 100.00% | 100.00% |
| 77 | 88.4263075 | **100.00%** | **100.00%** | **100.00%** | 50.00% | 50.00% | 50.00% |
| 82 | 56.0501649 | 100.00% | 100.00% | 100.00% | 100.00% | 100.00% | 100.00% |
| 84 | 22.8809663 | **71.43%** | **83.33%** | **76.92%** | 80.00% | 66.67% | 72.73% |
| 85 | 68.5711141 | **100.00%** | **100.00%** | **100.00%** | 50.00% | 100.00% | 66.67% |
| 87 | 49.7199117 | 80.00% | 100.00% | 88.89% | 80.00% | 100.00% | 88.89% |
| 88 | 81.8170171 | 100.00% | 100.00% | 100.00% | 100.00% | 100.00% | 100.00% |
| 89 | 14.206154 | 100.00% | 100.00% | 100.00% | 100.00% | 100.00% | 100.00% |
| 93 | 13.5899878 | 100.00% | 100.00% | 100.00% | 100.00% | 100.00% | 100.00% |
| 95 | 57.426799 | 100.00% | 100.00% | 100.00% | 100.00% | 100.00% | 100.00% |
| 96 | 11.8740355 | 100.00% | 100.00% | 100.00% | 100.00% | 100.00% | 100.00% |
| 98 | 61.3983121 | 100.00% | 100.00% | 100.00% | 100.00% | 100.00% | 100.00% |
| 123 | 54.5686799 | 100.00% | 100.00% | 100.00% | 100.00% | 100.00% | 100.00% |
| 125 | 29.4063831 | **100.00%** | **100.00%** | **100.00%** | 100.00% | 71.43% | 83.33% |
| 131 | 65.3736027 | 100.00% | 100.00% | 100.00% | 100.00% | 100.00% | 100.00% |
| 134 | 45.5904438 | 100.00% | 100.00% | 100.00% | 100.00% | 100.00% | 100.00% |
| 138 | 72.2330041 | 100.00% | 100.00% | 100.00% | 100.00% | 100.00% | 100.00% |
| 142 | 73.576066 | 100.00% | 100.00% | 100.00% | 100.00% | 100.00% | 100.00% |
| 143 | 24.0986572 | 100.00% | 100.00% | 100.00% | 100.00% | 100.00% | 100.00% |
| 180 | 10.3440649 | 100.00% | 100.00% | 100.00% | 100.00% | 100.00% | 100.00% |
| 181 | 84.1780428 | 100.00% | 100.00% | 100.00% | 100.00% | 100.00% | 100.00% |
| 183 | 81.9879789 | **100.00%** | **50.00%** | **66.67%** | 75.00% | 75.00% | 75.00% |
| 184 | 26.0136719 | 100.00% | 100.00% | 100.00% | 100.00% | 100.00% | 100.00% |
| 186 | 35.5071598 | 100.00% | 100.00% | 100.00% | 100.00% | 100.00% | 100.00% |
| 211 | 22.1367073 | 100.00% | 100.00% | 100.00% | 100.00% | 100.00% | 100.00% |
| 217 | 95.5176921 | 100.00% | 100.00% | 100.00% | 100.00% | 100.00% | 100.00% |
| 222 | 65.4059773 | 100.00% | 100.00% | 100.00% | 100.00% | 100.00% | 100.00% |
| 226 | 43.1260097 | 100.00% | 100.00% | 100.00% | 100.00% | 100.00% | 100.00% |
| 227 | 44.3401323 | **100.00%** | **100.00%** | **100.00%** | 100.00% | 75.00% | 85.71% |
| 229 | 16.0993804 | 100.00% | 100.00% | 100.00% | 100.00% | 100.00% | 100.00% |
| 231 | 60.5148326 | **83.33%** | **100.00%** | **90.91%** | 100.00% | 100.00% | 100.00% |
| 232 | 17.4316837 | 100.00% | 100.00% | 100.00% | 100.00% | 100.00% | 100.00% |